# PERFORMANCE EVALUATION OF MACHINE LEARNING TECHNIQUES FOR DOS DETECTION IN WIRELESS SENSOR NETWORK

Lama Alsulaiman and Saad Al-Ahmadi

Department of Computer Science, King Saud University, Riyadh, Saudi Arabia

## ABSTRACT

*The nature of Wireless Sensor Networks (WSN) and the widespread of using WSN introduce many security threats and attacks. An effective Intrusion Detection System (IDS) should be used to detect attacks. Detecting such an attack is challenging, especially the detection of Denial of Service (DoS) attacks. Machine learning classification techniques have been used as an approach for DoS detection. This paper conducted an experiment using Waikato Environment for Knowledge Analysis (WEKA)to evaluate the efficiency of five machine learning algorithms for detecting flooding, grayhole, blackhole, and scheduling at DoS attacks in WSNs. The evaluation is based on a dataset, called WSN-DS. The results showed that the random forest classifier outperforms the other classifiers with an accuracy of 99.72%.*

## KEYWORDS

*Wireless Sensor Networks, Machine Learning, Denial of Service*

## 1. INTRODUCTION

Wireless Sensor Networks (WSN) is one of the important research topics in computer science. The WSN is a preferred solution for many applications in different fields such as medical and health care, telecommunications, moreover, WSN can be used during natural disasters to detect flooding, volcanoes, or earthquakes[1]. Due to the widespread use of WSN, it introduces many security threats. The WSN is vulnerable to different kinds of attacks due to a variety of constraints such as limited processing, battery power, storage space. Denial of Service attack (DoS) is the most common type of attack that can affect WSN. DoS attacks can harm a network service by sending a huge number of fake requests to overwhelm the network resources to a point that legitimate traffic is prevented from accessing the network. The intrusion detection system (IDS) should be used to ensure the WSN is secure.

Machine learning (ML) techniques are considered to be one of the prominent methods that could be used with IDSs to improve their ability to identify and recognize attackers. The ML classification method has been used for DoS detection in WSN. The goal of this research is to investigate several classification models Naïve Bayes, Neural networks, Support Vector Machine, Decision tree, and Random forest to assess which model classifies the data set the best. These techniques are compared using different comparison measures such as accuracy, precision, recall and evaluated on a specialized WSN dataset named WSN-DS [2] containing normal and multiple attack scenarios to verify their efficacy in the detection of DoS attacks.

The rest of this research is organized as follows: in section 2 a brief background about the used machine learning algorithms and DoS attacks in WSN, followed by section 3 presents the literature review, evaluation, and experimental results discussed in section 4. Finally, the conclusion is presented in section 5.

## 2. LITERATURE REVIEW

Several studies have explored detecting and classifying security attacks in general and WSN attacks in particular. This section briefly reviews some related work about IDSs in WSNs.

A comprehensive literature review was presented in [3], discussing the strategies, algorithms, and applications of ML in WSNs. It also discussed some challenges such as energy awareness, query processing, event detection, security, and quality of service. However, this work undertook only a qualitative evaluation, and quantitative results are missing.

In [2], Almomani et al. built a dataset for WSNs containing four types of DoS attacks: blackhole, grayhole, flooding, and scheduling. The collected dataset is called WSN-DS and is intended to help researchers detect DoS attacks in WSNs. Almomani et al. used this dataset to train an artificial NN (ANN) to detect and classify DoS attack types. The experimental results show that different types of DoS attacks were detected with higher accuracy. The best results were achieved with one hidden layer.

Gonduz et al. [4] presented a survey of ML solutions for detecting DoS attacks. The DoS variations were reviewed at each layer of the TCP/IP protocol stack and focused on network layer attacks.

Belavagi and Muniyal[5] presented a performance evaluation of supervised ML algorithms for intrusion detection in WSNs using the NSL-KDD dataset. However, the evaluation focused only on the ability to detect intrusions.

The authors in [6] used ML algorithms to improve anomaly detection in a WSN. The experiment used a medical dataset from PhysioNet. The results showed that the J48 algorithm performed best for classification and that KNN was best for regression tasks.

The authors in [7] presented a lightweight detection system using logistic regression that studied the behavior of a large number of nodes under jamming and blackhole attacks. Various parameters were considered in this experiment, such as traffic intensity, transmission power, and attacker location. The authors also considered different topologies in WSNs, such as central data collection and meshed multi-hop networks.

Nancy et al. [8] proposed a dynamic recursive feature selection algorithm to select an optimal number of features from a dataset along with a fuzzy temporal DT algorithm. They extended the DT algorithm and integrated it with NNs to effectively detect attacks. The KDD Cup dataset was used in the experiment, which demonstrated the effectiveness of the proposed approach.

A multi-level IDS was proposed in [9] to secure WSNs using danger theory, which is derived from basic principles of artificial immune technology. The system monitors WSN parameters such as energy, data volume, and data transmission frequency to obtain output based on the weights and concentrations of these parameters. The robustness of the network is enhanced by cooperation between nodes to identify intrusions; it provides higher detection rates and lowers false detection rates and energy consumption. However, the detection rate is decreased, and the energy consumption is increased whenever the number of nodes increases.

[10] proposed a framework called Scale-HybridIDS-AlertNet that used hybrid deep NNs (DNNs) to monitor network traffic and protect a network from attacks. The DNN model was applied to different datasets, such as UNSW-NB15, Kyoto, WSN-DS, and CICIDS 2017. Their approach

performed better than other approaches using traditional ML classifiers in terms of detection accuracy and false alarm rate. The problem with this approach, however, is that it requires a large computational cost when building complex DNNs.

The authors in [11] proposed a new feature selection method called the conditional random field and linear correlation coefficient-based feature selection algorithm. Detection accuracy was increased by selecting some features and classifying them using a convolutional NN. The proposed system consists of six components: a KDD dataset, a user interface module, a feature selection module, a classification module, a decision manager, a rule manager, and a rule base. The KDD 99 Cup dataset was used to evaluate the system, which achieved 98.88% overall detection accuracy.

In [12], Ioannou et al. proposed an IDS called mIDS. This system is based on binary logistic regression, which classifies the activities of sensors as benign or malicious. A routing layer attack was used to evaluate the system and showed that mIDS could detect malicious activity in the 88−100% range with 91% accuracy.In the following table 1, summarize the literature review.

Table 1.  Summary of the literature review

| Ref | ML Technique | Dataset | Evaluation Metrics | Simulation Environment |
|---|---|---|---|---|
| 3 | Literature review from 2002–2013 of machine learning methods in WSN | - | - | Review |
| 2 | ANN | WSN-DS | TPR, TNR, FPR, FNR, Accuracy, Precision, Root Mean Square Error | NS-2 WEKA |
| 4 | Review of different studies used ML | - | - | Review |
| 5 | Logistic Regression Gaussian Naive Bayes Support Vector Machine Random Forest | NSL-KDD | Precision, Recall, F1-Score, Accuracy, and ROC | Not mentioned |
| 6 | J48, Random Forests, k-Nearest Neighbors, Linear Regression, Additive Regression, Decision Stump | MIMIC | ROC, mean error, time | Not mentioned |
| 7 | Logistic regression | Not mentioned | Basic metrics, CTP specific metrics, and mesh network-specific metrics. | Not mentioned |
| 8 | fuzzy temporal decision tree, Neural Network | KDD cup | Precision, Recall, F−measure | NS2 |
| 9 | Artificial Immune | Not mentioned | Time, Energy Overhead, Probability of Detection | COOJA |
| 10 | Deep neural networks | KDDCup 99, NSL-KDD, UNSW-NB15, Kyoto, WSN-DS | Accuracy Precision F−measure, TPR, FPR, ROC | Not mentioned |
| 11 | Convolutional neural network | KDD 99 | Accuracy, Time, false alarm rate | Not mentioned |
| 12 | Binary Logistic Regression | Not mentioned | Accuracy | COOJA |

## 3. BACKGROUND

This section introduces preliminary information necessary for subsequent sections as well as some concepts related to ML classification techniques and DoS attacks on WSNs.

### 3.1. Machine Learning Classification Techniques

IDSs for WSNs are classified into four categories based on whether they employ signature-based, anomaly-based, specification-based, or hybrid technique[13].Signature-based techniquesare a common technique for detecting well-known attacks. This technique, which classifies traffic samples based on known patterns from the training dataset, is known for its high accuracy and low false-positive rate. While this method is effective, it also suffers from some drawbacks; for instance, it is less efficient at detecting an unknown type of attack, since it needs to have a signature for the training dataset. Classification algorithms depend on signature-based techniques since such algorithms use known classes for training. Five classification techniques are considered in this paper.

The NB classifier is a form of Bayesian classifier, a group of probabilistic classifiers based on Bayes' theorem. NB classifiers have the property ofclass-conditional independence,meaningthat the value of a given attribute is assumed to be independent of the values of other attributes in a given class. It assumes that, if a particular feature exists in a class,it is unrelated to the existence of any other feature [14].

NNs, which were developed in 1943, is one of the most successful and powerful types of ML algorithms. They can be used as pieces of different ML algorithms to handle complex data by translating them into a form that the computer can understand. The NN is a graph with a set of nodes called neurons that are connected by edges. The architecture of an NN consists of input and output neurons as well as hidden layers, which process data by taking input from the input layer and producing output to the output layer. AnNN is based on a learning algorithm that learns from training datasets to help to produce the correct output. After the network learns, it can process new, previously unseen inputs and return the correct results [14].

SVMs can be used for classification and regression purposes. An SVM is a supervised learning method that is most commonly used in classification problems. SVMsaim to find the hyperplane that best divides a given dataset into classes. The hyperplane refers to the separation between classes and is found by locating the support vectors (the data points near the hyperplane) and the margins (the distance between the hyperplane and the closest data point). The hyperplanes can be found using a margin that maximizes distance. In SVM models, the data items are plotted as points in n-dimensional space, and the classification process is then performed by locating the hyperplane that finds the classes [14].

A DT is an algorithm for making decisions. It is an acyclic graph used to solve a problem by presenting various available alternative solutions to that problem. The DT contains the root node, where it begins; the leaf nodes, which indicates the class; and the nodes in between, whichdecide which branch in the tree to go to next using a particular function. Reaching a decision requires following a path from the root to the leaf. There are classical univariate DTs such as J48, a divide-and-conquer algorithm that uses the training dataset to construct trees, then calculates the entropy to predict classes to classify the unknown samples [14].

RF is an ensemble learning method for classification and regression that creates multiple DTs from a randomly selected subset of the training set. It then decides the final class of the test object based on different DTs [14].

## 3.2. Denial of Service Attacks in Wireless Sensor Networks

There are several reasons for a DoS to occur, such as hardware failures, software bugs, environmental conditions, or even the combination of these factors [15].

Different types of DoS attacks have been identified. In this paper, we assess our experiments on four particular DoS types: blackhole, grayhole, flooding, and scheduling or Time-Division Multiple Access (TDMA) attacks. These attacks could target Low Energy Aware Cluster Hierarchy (LEACH) routing protocol. LEACH protocol is used to lower power consumption and due to its simplicity. A Cluster Head (CH) is used in LEACH which allows communication between group members and sinks. These attacks are implemented in the WSN-DS dataset provided by Almomani et al [2].

DoS attacks types are briefly described below:

- Black Hole attacks: the attacker plays the CH role. Then the attacker will keep dropping packets and not forwarding them to the sink node.
- Grayhole attacks: the attacker advertising itself as a CH for other nodes. After the forged CH receives packets it selectively or randomly discarding packets, therefore it will prevent the legitimate packets to be delivered.
- Flooding attacks: flooding attacks targeting LEACH protocol by sending a large number to the sensor to advertise itself as an advertising CH. This will lead to consuming energy, memory, and network traffic.
- Scheduling attack: It occurs during the setup phase when CHs set up TDMA schedules for the data transmission time slots. The attacker will change the behavior of the TDMA schedule from broadcast to unicast to assign all nodes the same time slot to send data. This will cause a packet collision which leads to data loss.

## 4. EXPERIMENT AND EVALUATION

This paper assess five different classification techniques using Waikato Environment for Knowledge Analysis (WEKA), a tool used to model ML algorithms and analyze data to summarize results in a useful manner [16]. WEKA was used in this research to build the NB, NN, SVM, DT, and RF classifiers. These algorithms were then applied to the specialized WSN-DN dataset, described below.

### 4.1. Dataset Description

WSN-DS dataset is collected by Almomani et al.[2] , they used the LEACH protocol to collect the dataset with features of WSN and with different attacking scenarios. It contains a total of 374661 records, and it has 19 attributes described in Table 2. Four different DoS attacks in the dataset: Blackhole (10049), Grayhole (14596), Flooding (3312), and TDMA (6638), and normal with the remaining 340066 records.

Table 2. Attribute Description

| # | Attribute name | Attribute description |
|---|---|---|
| 1 | Id | Unique ID to distinguish the sensor node |
| 2 | Time | Current simulation time of the node. |
| 3 | Is CH? | A flag to distinguish whether the node is CH with value 1 or normal node with value 0. |
| 4 | Who CH? | The ID of the CH in the current round. |
| 5 | Distance to CH | the distance between the node and its CH in the current round. |
| 6 | ADV_S | the number of advertise CH's broadcast messages sent to the nodes. |
| 7 | ADV_R | the number of advertising CH messages received from CHs |
| 8 | Join_S | the number of join request messages sent by the nodes to the CH |
| 9 | Join_R | the number of join request messages received by the CH from the nodes. |
| 10 | SCH_S | the number of advertise TDMA schedule broadcast messages sent to the nodes. |
| 11 | SCH_R | the number of TDMA schedule messages received from CHs. |
| 12 | Rank | the order of this node within the TDMA schedule. |
| 13 | DATA_S | the number of data packets sent from a sensor to its CH. |
| 14 | DATA_R | the number of data packets received from CH. |
| 15 | Data_Sent_To_BS | the number of data packets sent to the BS |
| 16 | dist_CH_To_BS | the distance between the CH and the BS. |
| 17 | send_code | the cluster sending code. |
| 18 | Consumed Energy | the amount of energy consumed in the previous round. |
| 19 | Attack Type | type of the node. It is a class of five possible values, Blackhole, Grayhole, Flooding, and TDMA, and normal, if the node is not an attacker |

## 4.2. Evaluation Metrics

Performance analysis of the intrusion detection algorithms can be done using the following measures: True Positive (TP), False Positive (FP), True Negative (TN), and False Negative (FN). From these measures, we can compute the following metrics which are used to assess the performance in DoS detection and to compare each classifier performance.

The detection accuracy represents the percentage of instances correctly classified

$$\text{Accuracy} = (TP+TN)/(TP+TN+FP+FN) \qquad (1)$$

The recall is the percentage of elements that have been classified correctly in each class as positive.

$$\text{Recall} = (TP)/(TP+FN) \qquad (2)$$

The Precision is the number of elements that have been classified correctly out of each class.

$$\text{Precision} = (TP)/(TP+FP) \qquad (3)$$

F1-measure is the average precision rate and recall rate.

$$\text{F1-measure} = (2 \times \text{Recall} \times \text{Precision})/(\text{Recall}+\text{Precision}) \qquad (4)$$

Also, we consider the timecomplexity which is the time taken to build the model.

## 4.3. Experimental Results and Discussion

The below presents a detailed comparative analysis of the five classification algorithms in terms of detection accuracy, recall, precision, $F_1$-measure, and time complexity. We used 10-fold cross-validation to conduct the experiment and to train the classifiers.

Figure 1summarizes the accuracy of the five classifiers. NB had the lowest accuracy at 95.35%, since it incorrectly classified 17,419 instances, whereas RF had the highest detection accuracy at 99.72%. J48 came second at 99.66%, and the NN achieved 98.57% detection accuracy. The SVM had an accuracy rate of 97.11%.

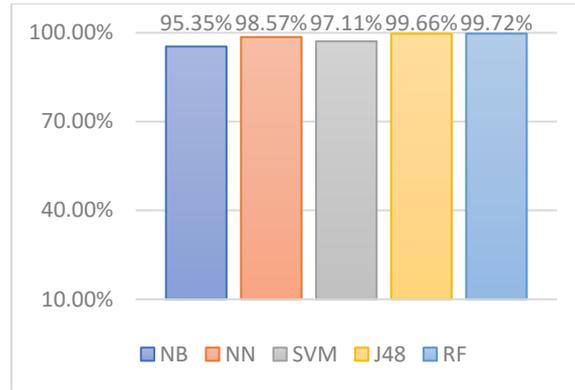

Figure 1. Detection Accuracy

Figure 2 summarizes the algorithms performance in terms of recall, precision, and $F_1$-measure. First, we discuss therecall results in terms of average recall for all data instances. J48 and RF had the highest average recall (0.997) whereasNB had the lowest recall (0.954). By type of attack,RF performed best in detecting blackhole and grayhole attacks, while NB was the best at detecting flooding attacks. On the other hand, the SVM was the worst at detecting grayhole attacks.

Regarding the precision results of each classifier, theclassifier with the lowest average precision was NB, and J48 and RF had the highest average precision (0.997). By type of attack,RF wasthe best at detecting blackhole and grayhole attacks,whereas NB was the worst at detecting flooding attacks.

Regarding the results for$F_1$-measure, J48 and RF obtainedthe highest average $F_1$-measure value (0.997),whereas the lowest value was 0.957 using NB.

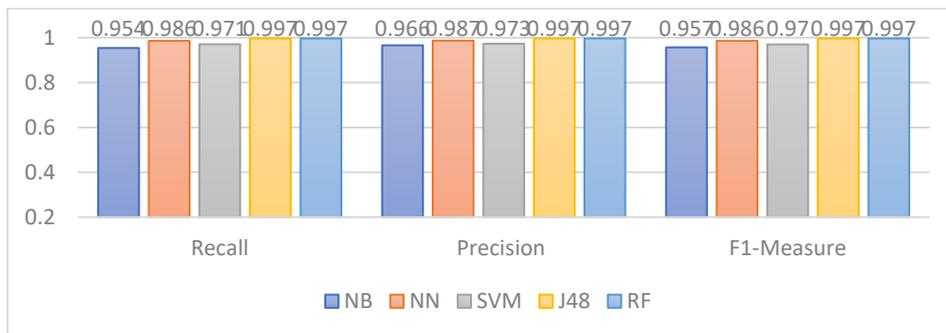

Figure 2. Performance in terms of Recall, Precision, and F1-Measure

To compare the performance of the classifiers, we also compared the time taken to build each model in Figure 3. NB was the fastest classifier to build at only 1.25 seconds. The NN took the longest amount of time to build at 647.06 seconds—equivalent to around 10 minutes—since it contains a large number of neurons, whose connectivity takes time.

Based on the previous results and considering all metrics, we conclude that the most accurate classifier is RFsince it demonstratedbetter results than other techniques and takes a reasonable time of amount to build. RF has the best accuracy in detecting attacks due to the selection of multiple trees and can more effectively handle a large volume of data. The worst technique to detect DoS attacks was NB; although it was the fastest technique, it also had the worst results compared to other techniques. The NB made someinaccurate assumptions because it depends on class-conditional independence and assumes that features are independent.

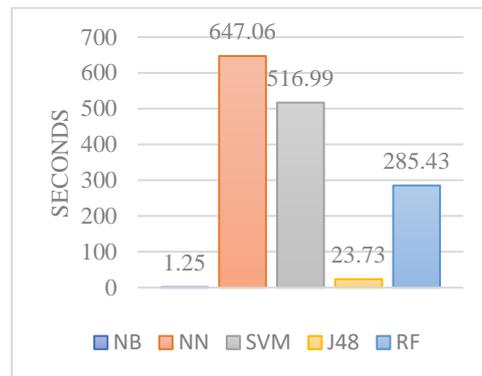

Figure 3. Time Complexity

## 5. CONCLUSION

In this work, we evaluatedML classification techniques for DoS detection in WSNs. A specialized dataset for WSNs called WSN-DS was used to compare the performance of ML techniques. The dataset was used to classify four types of DoS attacks: blackhole, grayhole, flooding, and TDMA. Different ML techniques were tested:NB, NN, SVM, DT, and RF. WEKA were used to test the performance of DoS detection on the WSN-DS dataset. We found that RF achieved better classification results than other techniques with a detection accuracy of 99.72%. In the future, this research will be extended to include other types of classifiers and ML techniques. It isalso possible for future research to consider other attack scenarios.

## AUTHORS

**Lama Alsulaiman** received her bachelor's degree in Computer Science from Imam Muhammad ibn Saud Islamic University. She is currently pursuing the masterdegree in the Computer Science program at King Saud University. Her research interests are mainly in the field of Computer Networks, Networks Security, Software-Defined Networks.

**Saad Al-Ahmadi** received the MS and Ph.D. degree in computer science from King Saud University, Saudi Arabia. He is an Associate Professor in the Department of Computer Science, King Saud University. Also, he serves as a part-time consultant in many public and private organizations. His current research interests include Cybersecurity, IoT, machine learning for healthcare, and future generation networks.

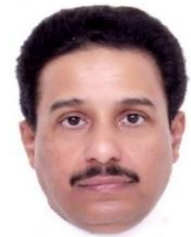